\documentclass{article}

\usepackage{arxiv}

\usepackage[utf8]{inputenc} 
\usepackage[T1]{fontenc}    
\usepackage{hyperref}       
\usepackage{url}            
\usepackage{booktabs}       
\usepackage{amsfonts}       
\usepackage{nicefrac}       
\usepackage{microtype}      
\usepackage{lipsum}		
\usepackage{graphicx}
\usepackage{natbib}
\usepackage{doi}

\title{forester: A Tree-Based AutoML Tool in R}

\author{Hubert Ruczyński\\
	Faculty of Mathematics and Information Science\\
 Warsaw University of Technology \\
	\texttt{hruczynski21@interia.pl} \\
	\And
	Anna Kozak \\
	Faculty of Mathematics and Information Science\\
 Warsaw University of Technology \\
	\texttt{anna.kozak@pw.edu.pl} \\
}

\renewcommand{\shorttitle}{\textit{arXiv} Template}

\hypersetup{
pdftitle={A template for the arxiv style},
pdfsubject={q-bio.NC, q-bio.QM},
pdfauthor={David S.~Hippocampus, Elias D.~Striatum},
pdfkeywords={First keyword, Second keyword, More},
}

\begin{document}
\maketitle

\begin{abstract}
The majority of automated machine learning (AutoML) solutions are developed in Python, however a large percentage of data scientists are associated with the R language. Unfortunately, there are limited R~solutions available. Moreover high entry level means they are not accessible to everyone, due to required knowledge about machine learning (ML). To fill this gap, we present the \textit{forester} package, which offers ease of use regardless of the user's proficiency in the area of machine learning.\\

\noindent The \textit{forester} is an open-source AutoML package implemented in R~designed for training high-quality tree-based models on tabular data. It fully supports binary and multiclass classification, regression, and partially survival analysis tasks. With just a few functions, the user is capable of detecting issues regarding the data quality, preparing the preprocessing pipeline, training and tuning tree-based models, evaluating the results, and creating the report for further analysis. 
\end{abstract}


\section{Motivation and significance}

Machine learning is being used more and more in the world around us. Every day, models are created to assist doctors \citep{oncology}, financiers \citep{credit_evaluation}, or tourists \citep{recommendation_system}. With the increasing demand for model building, research is being conducted on automatically developing tools to build artificial intelligence-based solutions. 

Many types of models are used in machine learning, ranging from decision rules to complex neural network structures modeling natural language (e.g. ChatGPT \citep{ChatGPT}). Viewing machine learning in terms of tabular data, we have a~wide range of models available, from decision trees \citep{quinlan:induction} and linear or logistic regression \citep{LogisticRegression} to random forests \citep{breiman2001random}, SVM \citep{SVM}, or neural networks \citep{mcculloch1943logical}. However, tree-based models are the most widely used; the main reason behind this is their high predictive efficiency. A simple decision tree model gives relatively satisfactory results, but using multiple trees to create a random forest allows significantly higher predictive power \citep{evaluation_models, trees_are_cool}.

Automating the process to build machine learning models can include many different components. The most common approach for building an automatic machine learning system is the preparation of models based on data entered by the user. This process can be extended in various directions for example, by automating the exploratory data analysis (EDA). Another essential element is the exploration of the search space of the model's hyperparameters, where we can choose from testing the values from a predefined grid, using random search, or employing the Bayesian optimization \citep{BayesOpt} or meta-learning \citep{metalearningmoremore, metalearningmore, woznica2022towards} approaches. After tuning the models, we can also automate the process of analyzing the results in the form of a~leaderboard, visualization, or even reporting system.


Packages for AutoML are prevalent in Python. The first AutoML solutions like Auto-WEKA \citep{thornton2013auto}, were followed by more advanced tools such as TPOT (Tree-Based Pipeline Optimization Tool) \citep{olson2016tpot}, auto-sklearn \citep{askl-1,askl-2}, mljar-supervised \citep{mljar}, and AutoGluon \citep{AutoGluon}. But in~R~\citep{R}, there are limited approaches \citep{mlr3, H2O}. 

One of the best of them is the \textit{H2O} package \citep{H2O}. It~is an open-source library that is an in-memory, distributed, fast, and scalable machine learning and predictive analytics platform that creates a ranked list of models easily exported for use in a~production environment. \textit{H2O}'s AutoML is also designed for more advanced users by providing a simple wrapper function that performs many modeling tasks. The main drawbacks of this framework are the limited evaluation options (no support for custom metrics), lack of a preprocessing module, and the necessity to run Java processes in the background, which terminates itself after interruption of code execution.



In this paper, we present the AutoML package written in R  to create models for binary and multiclass classification, regression, and survival analysis tasks on tabular data. The main goals of the package are making the tool easy to use, fully automating all the necessary steps inside the ML pipeline, and providing results that are easy to create, understand, and allow diagnostics of the models. 
The implementation of the \textit{forester} package can be found in our GitHub repository\footnote{\url{https://github.com/ModelOriented/forester}}. The software is open-source and contains comprehensive documentation with examples of use. 

\section{Software description}

\begin{figure}
    \centering
    \includegraphics[width=16cm]{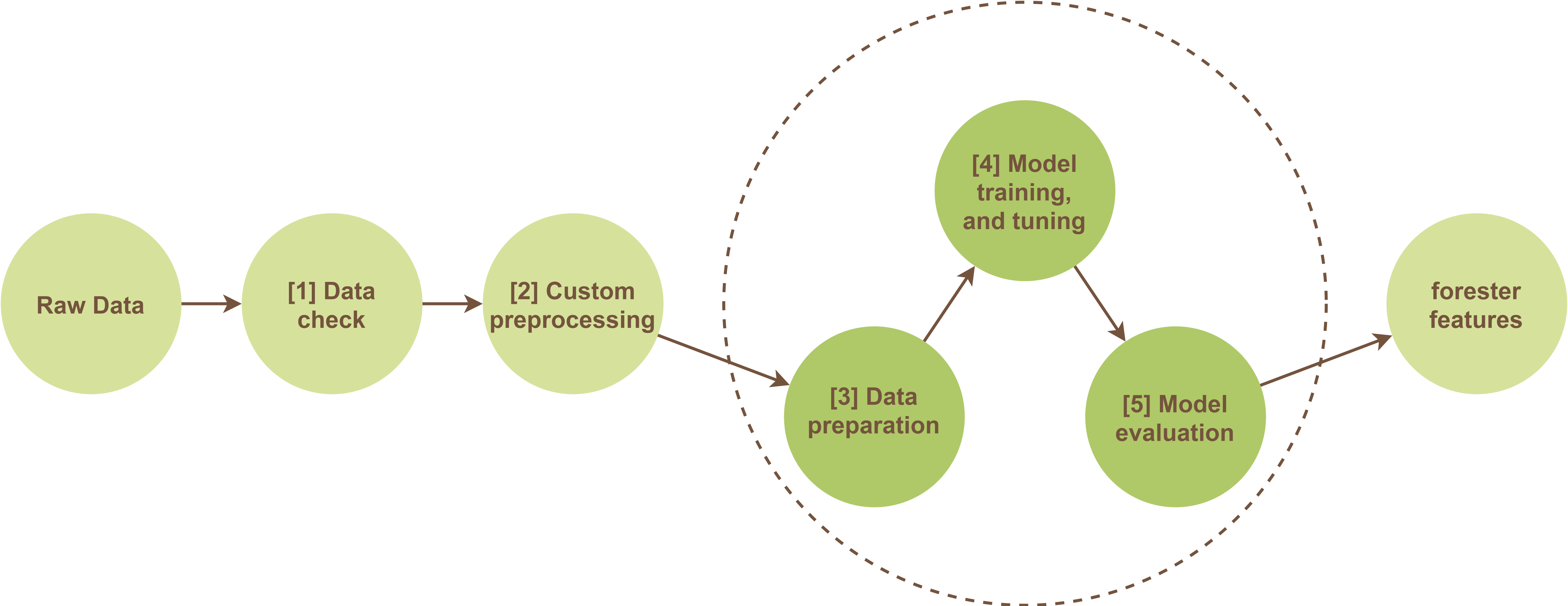}
    \caption{A diagram presenting the \textit{forester} pipeline. The numbers inside brackets indicate which paragraph from Subsection \ref{sec:software_arch} describe this part.}
    \label{fig:forester_pipeline}
\end{figure}

The \textit{forester} is an AutoML package automating the machine learning pipeline, starting from the data preparation, through model training, to the interpretation of the results. This way, we minimize the user's time performing basic and often repetitive activities related to the machine-learning process, such as data encoding, manual parameter choice, or evaluating the outcomes. Despite the high automation of the pipeline shown in Figure \ref{fig:forester_pipeline}, we expose multiple parameters that advanced data scientists can use to customize the model creation. The whole package relies on the five pillars.

\subsection{Software architecture}
\label{sec:software_arch}

\begin{enumerate}

\item \textbf{Data check} \\
The first one, called data check, concerns a data preprocessing phase. This module provides a list of possible issues existing within the used dataset. The analysis of~its results might help us choose the proper settings for the custom preprocessing module. 

\newpage

\item \textbf{Custom preprocessing} \\
The second pillar is custom preprocessing, being a high-level module that simplifies writing the code for a~well-suited preprocessing pipeline. Its usage is optional, although in most cases advisable. The~module covers three major parts of data quality enhancements: the removal of~corrupted features, the imputation of~missing data, and feature selection. 

The sub-module responsible for the deletion of irrelevant features uses up to 6 separate criteria that determine whether a column or row should be removed or not. It is possible to build various unique strategies based on the following removal options:
\begin{itemize}
    \item Duplicated columns - remove features that are identical,
    \item Id-like columns - remove features resembling IDs based on either a provided or a default list,
    \item Static columns - remove features exhibiting a high percentage (governed by user-defined threshold $k$) of identical values,
    \item Sparse columns - remove features with a low percentage (governed by user-defined threshold $l$) of non-empty values,
    \item Corrupted rows - remove observations with either a low percentage (governed by user-defined threshold $m$) of non-empty values or an empty target value,
    \item Highly correlated columns -  iteratively remove the minimal number of features exceeding a user-defined correlation threshold ($n$) to achieve the desired outcome of reducing multicollinearity.
\end{itemize}
Parameters $k, l, m, n$ are threshold parameters provided by the user, corresponding to the appropriate criterion, and its values.

The second sub-module determines which imputation method to use if the dataset contains any missing values. The user can choose from 4 algorithms. With the median-other, and median-frequency approaches the numeric features are imputed with median value, whereas the categorical ones with the 'other' string or the most frequent value. It is also possible to choose more advanced algorithms such as K-Nearest Neighbors (KNN), \citep{knn} from \textit{VIM} package \citep{VIM}, or Multivariate Imputation by Chained Equations (MICE) \citep{MICE} and determine their most important parameters.

The last part of the preprocessing module is the choice of feature selection method. The package offers four state-of-the-art (SOTA) algorithms, being Mutual Information (MI) based method \citep{MI} from \textit{varrank} package \citep{varrank}, Boruta \citep{BORUTA}, Monte Carlo Feature Selection (MCFS) \citep{MCFS}, and Variable Importance (VI) \citep{VI}. The interface of function enables the user to fine-tune the most important parameters of these methods. The~selection of algorithms covers various time-complexity, where MCFS and MI are relatively fast, whereas Boruta and VI are more time-consuming.


\item \textbf{Data preparation} \\
As the previous step is not obligatory, at the beginning of this stage the package conducts the most basic preprocessing if the user did not use custom preprocessing beforehand. It consists of steps where we remove static columns, impute missing values with KNN, and provide consistent labels for classification tasks. Moreover, every algorithm in the \textit{forester} package requires a different data format which is also prepared inside the main function.

\item \textbf{Available tasks} \\
The fourth and most important pillar of the \textit{forester} package is model training and tuning.
The package currently fully supports three tabular, machine learning tasks, namely binary and multiclass classification, and regression. Additionally, we partially support survival analysis tasks. The solution focuses on the tree-based model family, which has been shown to have high-quality performance for various tabular data tasks \citep{trees_are_cool}. We have limited the choice of engines to decision tree \citep{quinlan:induction} from \textit{partykit} \cite{partykit}, random forest \citep{breiman2001random} from \textit{ranger} \citep{ranger}, XGBoost \citep{xgboost}, LightGBM \citep{lightgbm}, and CatBoost \citep{catboost} for classification, and regression tasks. Furthermore, we handle the survival analysis with random survival forests \citep{RSF}, \citep{RSFR} from \textit{randomForestSRC} package \citep{rfsrc}. These models are trained with three approaches: using the default parameters, performing the random search algorithm within the predefined parameter space, and running an advanced Bayesian optimization \citep{BayesOpt} algorithm for fine-grained tuning.
\newpage
\item \textbf{Model evaluation} \\
The final component of the forester package is the automated evaluation of trained models. The package assesses performance using a comprehensive set of metrics tailored to different task types. Accuracy, area under the receiver operating characteristic curve (AUC ROC), F1 score, recall, precision, sensitivity, specificity, and balanced accuracy are used for the evaluation of the binary classification tasks. The multiclass classification task is evaluated with accuracy, micro and macro averaged precision, recall, and F1 score, as well as with weighted precision, recall, and F1 score. For the regression models, it calculates Mean Squared Error (MSE), Root Mean Squared Error (RMSE), Mean Absolute Error (MAE), $R^2$, and Mean Absolute Deviation (MAD). The survival analysis task is evaluated with the usage of Brier Score, and Concordance Index. Furthermore, the user can provide a custom metric and incorporate it in the \textit{forester} pipeline. The results are later presented as a ranked list sorted by the selected metric.
\end{enumerate}

\subsection{Software functionalities}

One of the most important goals for the \textit{forester} package is the convenience of use and helping the users to focus more on analyzing the results instead of writing the code. To obtain such a~user-friendly environment, the \textit{forester} offers plenty of additional features useful for data scientists.

\begin{enumerate}

\item \textbf{Model explanations}

In recent years, interpretable machine learning has become a significant ML topic. The tools providing interpretability such as \textit{DALEX} \citep{DALEX} or \textit{iml} \citep{iml} allow data scientists to explain how the models they create work, making it easier to detect their misbehavior. To support using explainable methods for the models trained by the \textit{forester}, we have created a wrapper for the \textit{DALEX} explainer compatible with our package. This way, the user can easily create various explanations for the trained models.

\item \textbf{Saving the outcomes}

Another crucial feature is the save function, which lets the user save the training output. Returned \textit{forester} object contains lots of information, such as preprocessed dataset, split datasets, split indexes, ranked lists for training, testing, and validation datasets, the predictions of the model, and much more. The abundance of objects makes it incredibly important to save the outcomes after the time-consuming training process.

\item \textbf{Model selection}

The results of the train function can take lots of space, especially if the user includes all engines, and calculates plenty of random search models. Thus, we offer model selection function, which enables simple selection of proper output elements, based on the user-chosen models.

\item \textbf{Automated report}

Our solution offers an automatically generated report that helps users quickly and easily analyze the training results. The main goal of this feature is to ensure that every user is able to easily assess the quality of the trained models. The report consists of basic information about the dataset, a data check report, a ranked list of the best ten (by default) models, and visualizations concerning model quality.


\end{enumerate}

\newpage
\subsection{Sample code snippets analysis}
In order to show the \textit{foresters} ease of usage we present the exemplary snippets for the most important features. The~code chunks from Figures \ref{fig:train}, \ref{fig:custom_preprocessing}, \ref{fig:extensive_features} present all possible methods parameters, which are described in the packages documentation.

\subsubsection{Training function}

\begin{figure} 
    \centering
    \includegraphics[height=10cm]{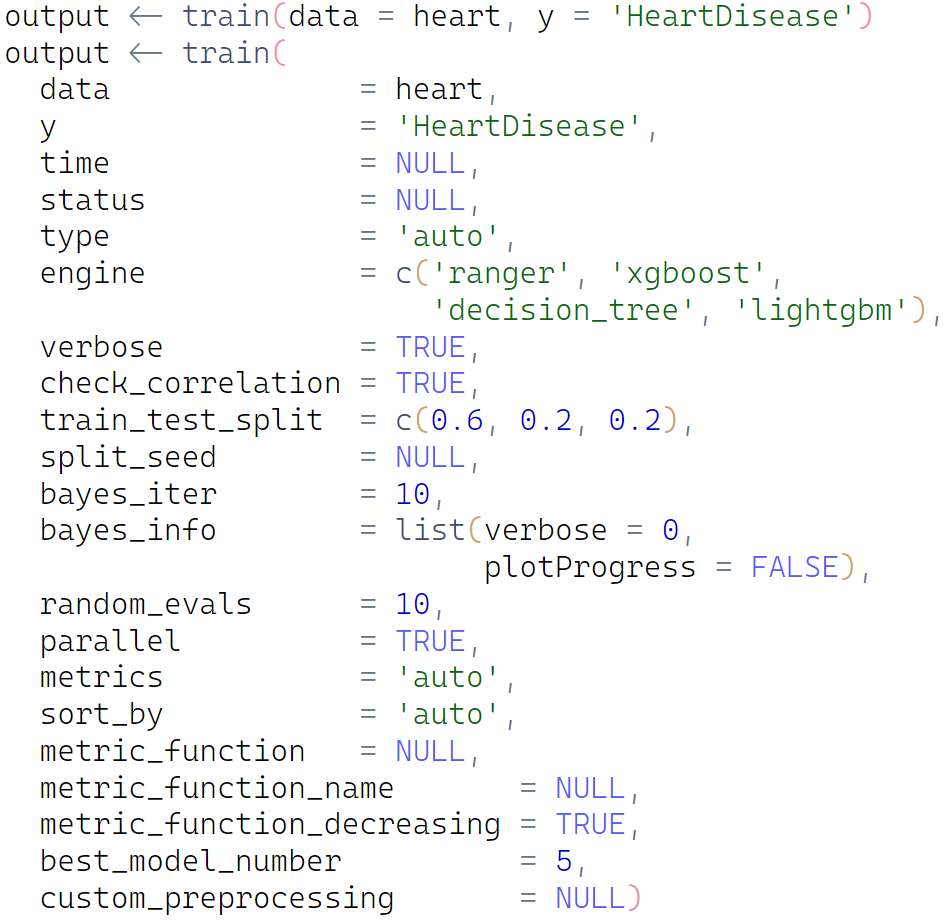}
    \caption{Training models with the \textit{forester} package.}
    \label{fig:train}
\end{figure}

The \textit{forester}'s main \texttt{train()} function, presented in Figure \ref{fig:train} runs the entire AutoML pipeline, including the data preparation, model training, and evaluation. To keep the package as simple as possible, the function requires only the dataset and target column name (first line); however, to keep the tool versatile, there are lots of custom parameters for more advanced users. 

\subsubsection{Custom preprocessing}

\begin{figure} 
    \centering
    \includegraphics[height=12cm]{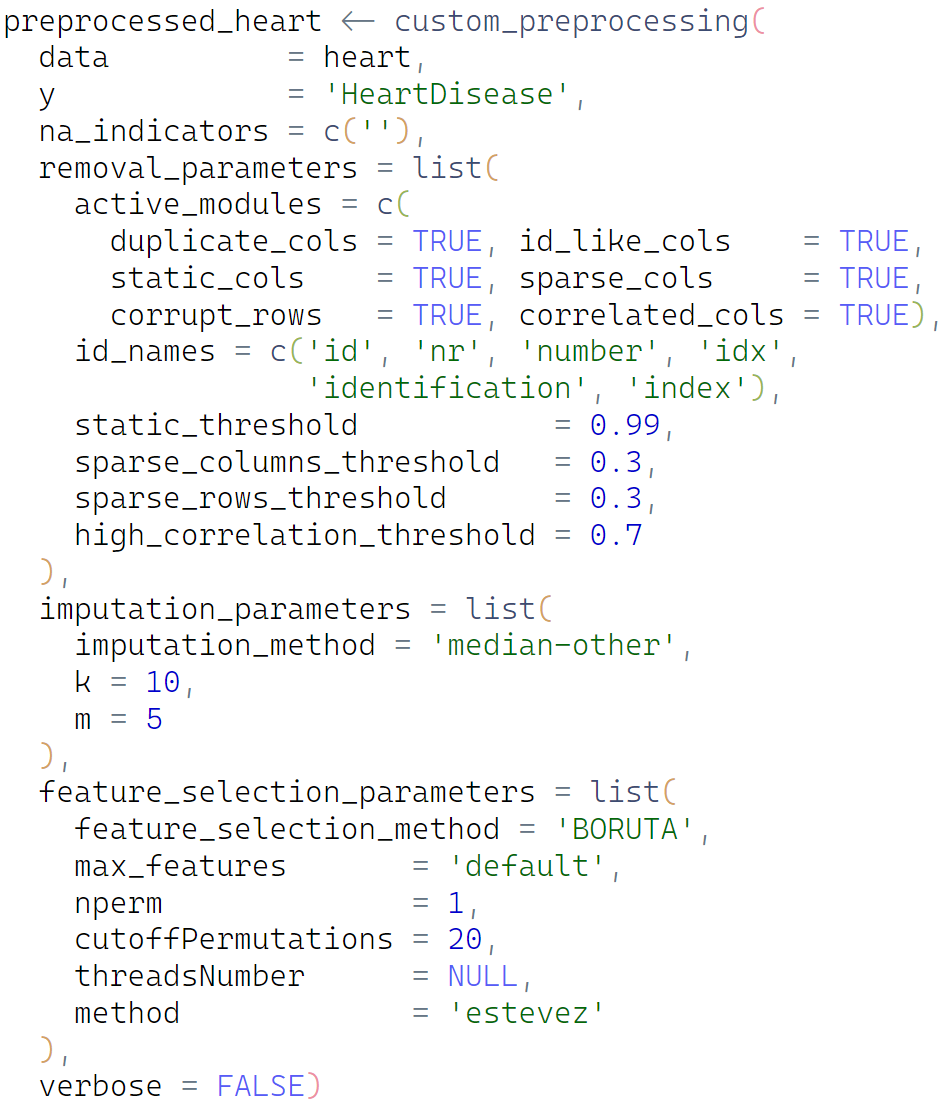}
    \caption{Preprocessing data with \texttt{custom\_preprocessing()} function. Its design underlines the division into 3~sub-modules, namely: parameters removal, data imputation, and feature selection.}
    \label{fig:custom_preprocessing}
\end{figure}

The most important extension to the aforementioned \texttt{train()} module is the \texttt{custom\_preprocessing()} function. It enables the user to easily design an automated and fine-grained preprocessing pipeline for the particular dataset. Its~interface is presented in Figure \ref{fig:custom_preprocessing}. 

\subsubsection{Extensive features}

\begin{figure}[!ht] 
    \centering
    \includegraphics[height=4.4cm]{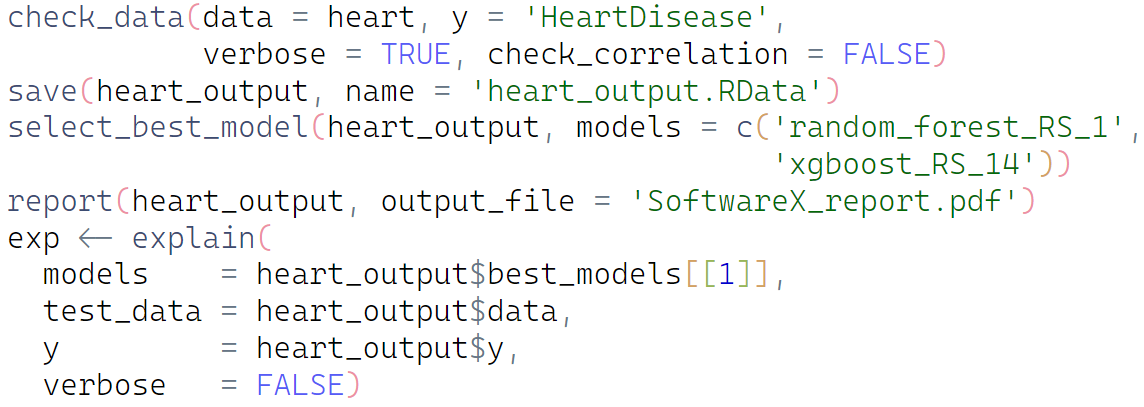}
    \caption{The interfaces of extensive functions of the \textit{forester}, which include data quality check, saving the outcomes, creating a training report, and explaining the best models.}
    \label{fig:extensive_features}
\end{figure}

The user can also utilize additional functions presented on Figure \ref{fig:extensive_features}, which are helpful during the modeling process. The \texttt{check\_data()} enables printing a data check report outside of the \texttt{train()} function. The \texttt{save()} function lets us save the outcome of the training process, \texttt{select\_models()} limits the train output, the \texttt{report()} creates a training report, whereas \texttt{explain()} creates \textit{DALEX} explainer for interpretable ML. 

\newpage
\clearpage
\section{Illustrative examples}

To better understand the possibilities, and comfort of using the \textit{forester} we present a short use case scenario in Figure \ref{fig:Use_case}, whose results are available in supplementary material\footnote{\url{https://github.com/ModelOriented/forester/blob/main/docs/articles/SoftwareX_paper/SoftwareX_report.pdf}}. We use the Heart Failure Prediction Dataset \citep{fedesoriano}, describing the task of binary classification, where we predict whether the person is likely to suffer from any heart disease. The dataset was created by combining different datasets already available independently but not combined before. 

We begin with loading the \textit{forester}, and the dataset (lines 1-2). Afterward, we run the \texttt{check\_data()} function (line 3), which shows us that despite a bunch of numeric outliers, the dataset does not have any major issues, so we can omit \texttt{custom\_preprocessing()}. The next step is model training, where we only have to provide the most necessary information and hyperparameter tuning settings (lines 4-9). Eventually, we take a closer look at the obtained results on the validation dataset (unseen in both training, and tuning, line 10), and generate the training report (line 11).

\begin{figure} 
    \centering
    \includegraphics[height=4.4cm]{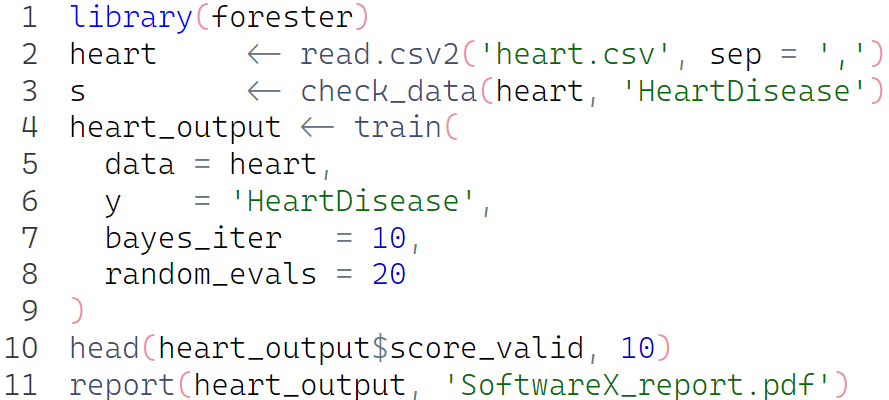}
    \caption{The use case scenario, presents the code required to conduct the whole ML process for heart disease detection task.}
    \label{fig:Use_case}
\end{figure}

\begin{figure}
    \centering
    \includegraphics[width=12cm]{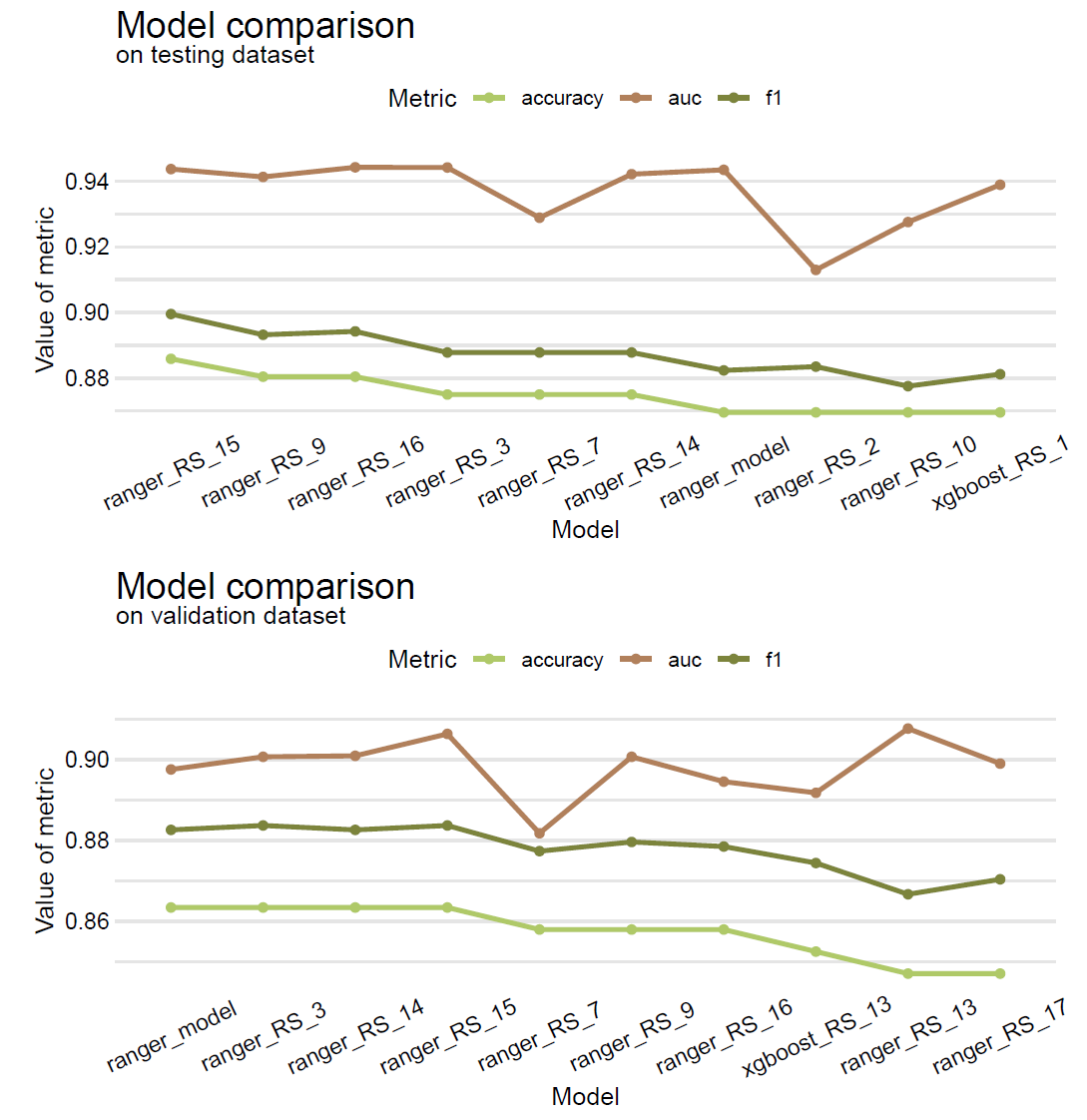}
    \caption{The comparison of \textit{forester} models performance, divided by type of the evaluation dataset.}
    \label{fig:model_comparison}
\end{figure}

If we analyze the model comparison section from a report, presented in Figure \ref{fig:model_comparison}, we can see that the outcomes achieved on testing, and validation datasets differ significantly. By default, the ranked list is sorted by accuracy, similar to the bottom plot. Thus ranger\_model is deemed the best one. 

\section{Impact}

The \textit{forester} is an answer to the omnipresent need for supplementing researchers' studies from different areas. The~tool's interface is very simple, clean, and enables the user to train various models from different R libraries with a single function. The package not only handles the model training, but it automatically conducts data preprocessing and preparation stages. Even more importantly it is capable of providing meaningful, and well-explained report, which describes the outcomes of trained models, enriched by the elements of interpretability. 

The \textit{forester} is perfect for beginners, as they can easily analyze some tasks with just a few lines of code. This~characteristic is extremely fruitful in the case of scientists from other fields. Even ML experts can benefit from it, because of the quick, and simple creation of baseline models, which lets them understand a new task quicker than usual. Furthermore, they can also use the data check, or custom preprocessing modules without the need to explicitly use the tool for model training. 

Additionally, the \textit{forester}'s GitHub repository already has a rich development history, and the tool is improved regularly to ensure the satisfaction of its users. It is reflected by over 100 stars on the website, which represents the community gathered around the package. It is also visible in the Issues history, with almost 100 entries, that the user's community is relatively active.

The development team is also focused on their work, as they popularize the solution by presenting the tool on international conferences (ML in PL, COSEAL, AutoML Conference), and conducting studies with the usage of the package. The most recent research topic revolves around the validation of a common saying that 'tree-based models do not require data preprocessing'. We are vividly exploring this area and plan to publish a paper on the matter as no studies confirm the aforementioned statement. Our tool also has been used in external research studies \citep{cavus2024experimental}.

\section{Conclusions}

This paper presents an R package for AutoML, creating models for binary, multiclass classification, and regression tasks conducted on tabular data, as well as partial support for survival analysis task. Our solution addresses the needs we have observed in AutoML tools in various programming languages. The main goals of the package are to keep the package stable and easy to use, to automate all the necessary steps inside the ML pipeline, and to provide results that are easy to create, understand, and allow for diagnostics of the models. To achieve these results, we have focused only on the best representatives from the family of tree-based models that show superiority over other methods on tabular data. Furthermore, we provide additional functions that allow the user to save the models, create explanations, and create a report describing the learning process and explaining the developed models.

\section*{Acknowledgments}

We would like to thank Adrianna Grudzień and Patryk Słowakiewicz for their early development work on the \textit{forester} package, and Przemysław Biecek for being one of supervisors. The custom preprocessing module, and survival analysis task were parts of the study financed from the competition CyberSummer@WUT-3 organized by Centrum Badawcze POB Cyberbezpieczeństwo i Analiza Danych, (CB POB Cyber\&DS) at Warsaw University of Technology.

\newpage

\bibliographystyle{apalike} 
\bibliography{references}

\end{document}